\documentclass[conference]{IEEEtran}
\IEEEoverridecommandlockouts

\usepackage{hyperref}
\usepackage{cite}
\usepackage{amsmath,amssymb,amsfonts}
\usepackage[numbers]{natbib}
\usepackage{graphicx}
\usepackage{textcomp}
\usepackage{xcolor}
\usepackage{algorithm}
\usepackage{algpseudocode}
\usepackage{booktabs}

\usepackage{tikz}
\usepackage{graphicx} 

\def\BibTeX{{\rm B\kern-.05em{\sc i\kern-.025em b}\kern-.08em
    T\kern-.1667em\lower.7ex\hbox{E}\kern-.125emX}}
\begin{document}

\title{Towards a Spatiotemporal Fusion Approach to Precipitation Nowcasting\\
}

\author{\IEEEauthorblockN{Felipe Curcio}
\IEEEauthorblockA{\textit{Cefet/RJ}\\
Rio de Janeiro, Brazil \\
0009-0004-4063-1613}
\and
\IEEEauthorblockN{Pedro Castro}
\IEEEauthorblockA{\textit{Cefet/RJ}\\
Rio de Janeiro, Brazil \\
0009-0001-4324-2910}
\and
\IEEEauthorblockN{Augusto Fonseca}
\IEEEauthorblockA{\textit{Cefet/RJ}\\
Rio de Janeiro, Brazil \\
0000-0003-1480-5814}
\and
\IEEEauthorblockN{Rafaela Castro}
\IEEEauthorblockA{\textit{SERPRO}\\
Brasília, DF \\
0000-0002-2136-6939}
\and
\IEEEauthorblockN{Raquel Franco}
\IEEEauthorblockA{\textit{AlertaRio}\\
Rio de Janeiro, RJ \\
0000-0003-0801-8363}
\and
\IEEEauthorblockN{Eduardo Ogasawara}
\IEEEauthorblockA{\textit{Cefet/RJ}\\
Rio de Janeiro, Brazil \\
eogasawara@ieee.org}
\and
\IEEEauthorblockN{Victor Stepanenko}
\IEEEauthorblockA{\textit{Lomonosov Moscow State University}\\
Moscow, Russia \\
v.stepanenko@rcc.msu.ru}
\and
\IEEEauthorblockN{Fabio Porto}
\IEEEauthorblockA{\textit{LNCC} \\
Petrópolis, Brazil \\
fporto@lncc.br}
\and
\IEEEauthorblockN{Mariza Ferro}
\IEEEauthorblockA{\textit{UFF} \\
Niterói, Brazil \\
mariza@ic.uff.br}
\and
\IEEEauthorblockN{Eduardo Bezerra}
\IEEEauthorblockA{\textit{Cefet/RJ}\\
Rio de Janeiro, Brazil \\
ebezerra@cefet-rj.br}
}

\maketitle

\begin{abstract}
With the increasing availability of meteorological data from various sensors, numerical models and reanalysis products, the need for efficient data integration methods has become paramount for improving weather forecasts and hydrometeorological studies. In this work, we propose a data fusion approach for precipitation nowcasting by integrating data from meteorological and rain gauge stations in Rio de Janeiro metropolitan area with ERA5 reanalysis data and GFS numerical weather prediction. We employ the spatiotemporal deep learning architecture called STConvS2S, leveraging a structured dataset covering a 9 x 11 grid. The study spans from January 2011 to October 2024, and we evaluate the impact of integrating three surface station systems. Among the tested configurations, the fusion-based model achieves an F1-score of 0.2033 for forecasting heavy precipitation events (greater than 25 mm/h) at a one-hour lead time. Additionally, we present an ablation study to assess the contribution of each station network and propose a refined inference strategy for precipitation nowcasting, integrating the GFS numerical weather prediction (NWP) data with in-situ observations.
\end{abstract}


\section{Introduction}
\label{sec:introduction}

Precipitation nowcasting (or very short-range forecasting~\cite{ahrens2019meteorology}) involves predicting rainfall within a six-hour lead time. This challenging task necessitates the integration of high-resolution temporal and spatial observations from diverse sources, including radars, satellites, lightning detection networks, surface stations, wind profilers, and radiosondes. Objective analysis techniques are then employed to synthesize these disparate measurements into a coherent, gridded spatial map for precipitation nowcasting~\cite{wmo_nowcasting}. Accurate precipitation forecasting is critical for mitigating natural disasters, such as floods, landslides, and droughts, and supports informed decision-making across sectors including agriculture, transportation, energy, and public health~\cite{Browning1982}.

Recent advancements in machine learning, particularly deep learning, have demonstrated significant potential in geoscientific applications, including precipitation nowcasting. Leveraging convolutional and recurrent architectures, deep learning models are well-suited to handle multidimensional, time-evolving structures inherent in geophysical phenomena, such as convective precipitation, which dominates tropical rainfall patterns. These methods facilitate the extraction of meaningful representations from complex datasets, addressing critical challenges in geoscientific and climate research~\cite{Reichstein2019195}.  

This study addresses two problems inherent in precipitation nowcasting through deep learning models. First, gauge station data is irregularly distributed, while most deep learning models work on gridded data. Second, precipitation data is highly imbalanced (where non-rain events overwhelmingly outnumber rain events), which makes it difficult to produce good forecasts for extreme precipitation values~\cite{KO2022105072}. To tackle these problems, we capitalize on the spatial resolution, uniformity, and completeness of both ERA5 reanalysis~\cite{era5} and Global Forecasting System (GFS) numerical weather prediction data~\cite{noaa_gfs}, and devise an algorithm to integrate gauge station observations. By integrating multiple surface stations, we apply data fusion techniques, which is the process of fusing multiple records representing the same object into a single, consistent, and more complete representation~\cite{bleiholder2009}. This allows us to overcome the limitations of individual data sources and leverage measurements of common and rare precipitation events from the available stations in the region of interest. The algorithm derives a distribution that emphasizes extreme precipitation events to enhance model training for the prediction of rare but impactful precipitation events. 

To validate our approach, we leverage on STConvS2S, a deep learning architecture for spatiotemporal weather forecasting~\cite{CASTRO2021285}. We build several dataset versions by considering different surface station combinations, and experimentally compare the predictive performance of the corresponding models. We find that integrating multiple surface stations enhances data completeness, particularly for capturing rare extreme precipitation events. Experimental results show that while our model correctly classified only 2.04\% of extreme events at a 1-hour lead time, it recognized 93.88\% of them as significant precipitation (either ``Moderate'' or ``Heavy''). This represents a meaningful improvement over Polifke's~\cite{polifke} study, where AlertaRio meteorologists misclassified the majority (63\%) of extreme events as ``No Rain'', thereby missing these critical precipitation events.

The remainder of this paper is structured as follows. Section~\ref{sec:related_work} presents related work on fusing data for precipitation nowcasting. Section~\ref{sec:methods} details our proposed approach for the construction of spatiotemporal series data. Section~\ref{sec:experiments} presents an ablation study, evaluating the impact of different surface station combinations, and provides an exploratory analysis and some sanity checks. We also present and evaluate the performance of the data fusion approach using STConvS2S for precipitation nowcasting. Finally, Section~\ref{sec:conclusions} summarizes our contributions and main findings, and outlines directions for future work.

\section{Related Work}
\label{sec:related_work}

The fusion of multiple data sources for precipitation nowcasting has been widely explored in recent studies, with approaches leveraging satellite data, radar observations, reanalysis products, and in-situ measurements.

Tosiri et al. \cite{tosiri2021} perform data fusion of radar observations from Weathernews Inc. (WNI) and satellite-based precipitation estimates from the Global Satellite Mapping of Precipitation (GSMaP) to enhance precipitation data coverage and improve the efficiency of precipitation nowcasting. The authors employ the U-Net~\cite{unet} image segmentation model for this task. Their findings indicate that data integration enhances forecasting efficiency compared to using radar data alone and extends coverage in regions with limited precipitation observations.  

A complementary approach is presented by Srithagon et al. \cite{rainfall2021}, in which the authors integrate data from a network of 130 rain gauge stations in the metropolitan area of Bangkok, Thailand, for precipitation nowcasting. The forecasting model is designed for a target station, selecting its 30 nearest neighboring stations as predictors. The authors conclude that precipitation forecasting with a lead time of up to 120 minutes can be effectively performed using only rain gauge data and a Random Forest model. Additionally, they observe that the F1-score is higher for shorter lead times.  

Kim et al. \cite{KIM2024105529} utilize radar reflectivity data along with two ERA5 variables — 925 hPa divergence and total column water vapor — for precipitation nowcasting. They also propose a balanced loss function, MCSLoss, to address class imbalance in precipitation forecasting. Their study concludes that incorporating atmospheric variables related to rainfall improves precipitation predictability and that the MCSLoss function enhances forecasting performance. Furthermore, the authors suggest that future studies should investigate the impact of additional atmospheric variables from reanalysis models and the potential benefits of integrating numerical weather prediction models.  

Choi and Cha \cite{rainFusion} introduce RAIN-F, a dataset for precipitation forecasting, and apply it to the U-Net model~\cite{unet}. The dataset integrates radar data, surface observations from the Automatic Weather Station (AWS) and the Automatic Surface Observing System (ASOS), as well as satellite-based precipitation estimates from IMERG. The authors report F1-scores of 0.69 and 0.19 for precipitation rates exceeding 0.1 mm/h and 5 mm/h, respectively. Compared to the performance of 0.68 and 0.03 when using only radar data for the same precipitation thresholds, \cite{rainFusion} conclude that the RAIN-F dataset provides superior performance, particularly in regions of heavy rainfall.  

Unlike previous studies that focus predominantly on radar, satellite, or rain gauge data in isolation, our work integrates meteorological and rain gauge station data with ERA5 reanalysis and GFS simulations in a structured deep learning framework. By leveraging the STConvS2S architecture, we assess the impact of multi-source fusion on precipitation predictability, particularly for extreme precipitation events. Our findings build upon existing research by evaluating the contributions of different surface station networks and proposing a refined inference strategy that incorporates both observational and numerical weather prediction data.

\section{Spatiotemporal Fusion Approach}
\label{sec:methods}

In this Section, we present our spatiotemporal fusion approach for precipitation nowcasting. We provide an overview of our proposed solution in Section~\ref{sec:overview}, which integrates multiple surface station networks with gridded reanalysis data or numerical weather prediction. Section~\ref{sec:dataset:building} introduces our dataset building methodology, explaining how we combine precipitation measurements from various surface stations with ERA5 to focus the data distribution on extreme precipitation events within our region of interest. Section~\ref{sec:model_building} presents the model development and evaluation process, presenting the STConvS2S and the metrics used to measure model performance among different precipitation levels.

\subsection{Overview of the Proposed Solution}
\label{sec:overview}

\begin{figure}[htb]
    \centering
    \includegraphics[width=0.45\textwidth]{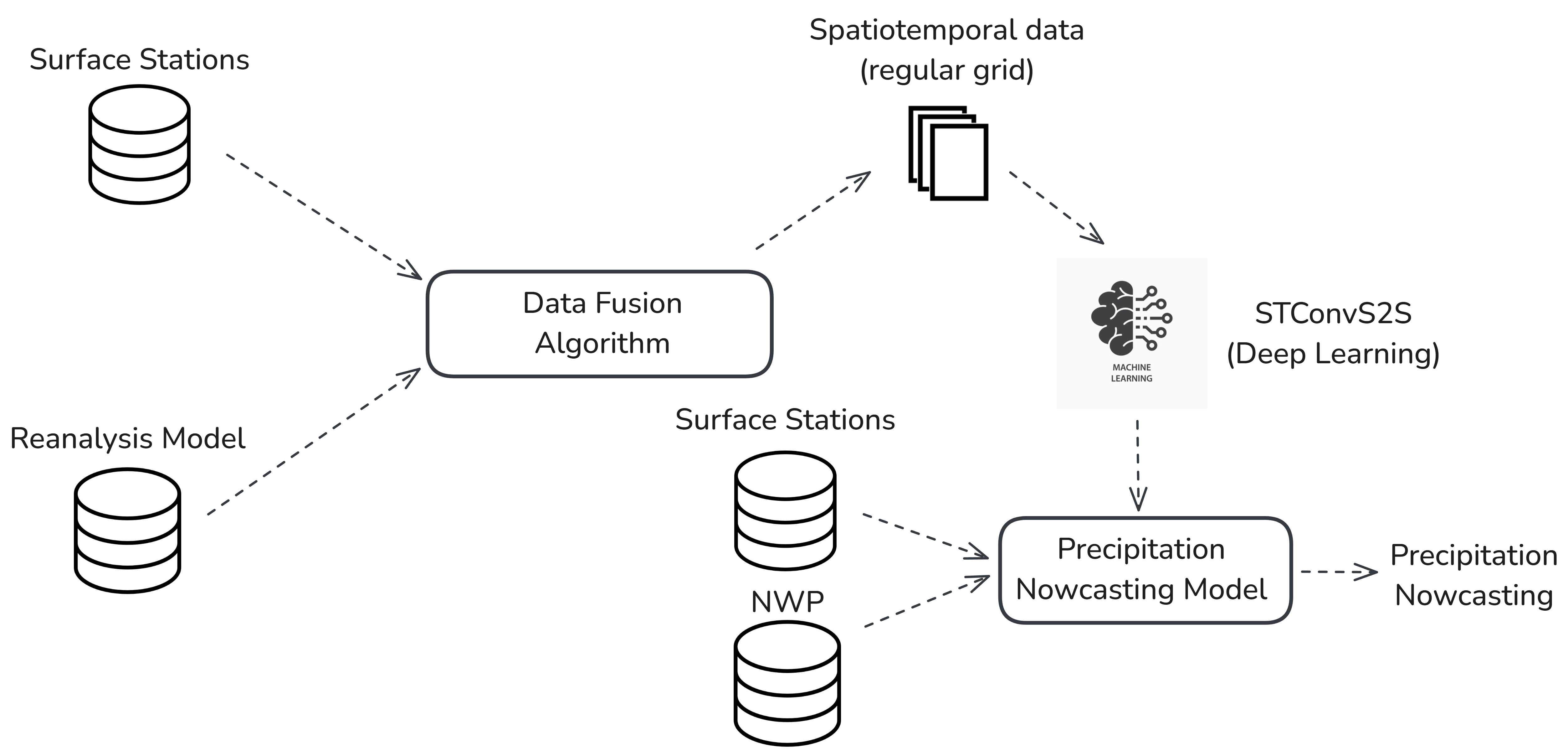}
    \caption{Pipeline for the proposed solution for precipitation nowcasting. The integration of surface station data with the ERA5 reanalysis model is performed to build datasets for model training and evaluation. At inference time, the integration of surface station data with NWP simulations is conducted to build data to feed the model.}
    \label{fig:proposed_solution}
\end{figure}

Our approach involves integrating multiple rain gauge stations with ERA5 reanalysis data during the training phase, and with numerical weather prediction (NWP) model forecasts during the inference phase. This enables the construction of uniform and regular grids with no missing data. Due to the five-day delay in the availability of ERA5 reanalysis data, during the inference phase, we integrate real-time surface station data with simulation data coming from a NWP model, namely GFS, as this data is available every 6 hours and provides predictions based on initial conditions. We employ the GFS model, which is compatible with the 0.25-degree spatial resolution of ERA5. Fig.~\ref{fig:proposed_solution} illustrates the pipeline for the proposed solution for spatiotemporal precipitation prediction.

\begin{figure}[htb]
    \centering
    \includegraphics[width=0.45\textwidth]{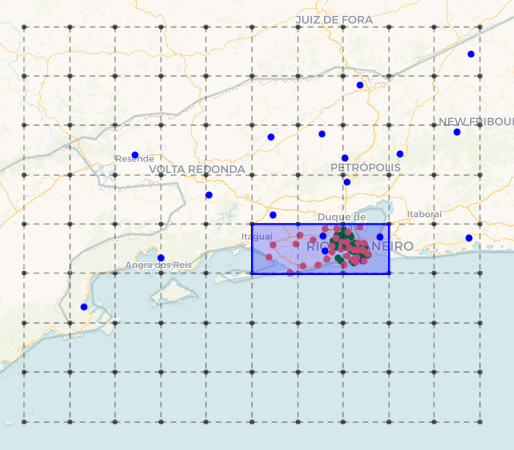}
    \caption{Regular ERA5 and GFS grid for the region of interest, highlighting the area where the model evaluation was conducted. The highest concentration of gauge stations corresponds to the three shaded cells, while the remaining grid cells are filled with values from the ERA5 reanalysis model. Station colors are: AlertaRio (red), Sirenes (green), and INMET (blue).}
    \label{fig:region_interest_highlight}
\end{figure}

In this work, we establish a rectangular region of interest, which is bounded to the north by latitude -21.6998, to the south by latitude -23.8019, to the east by longitude -42.3568, and to the west by longitude -45.0529, as illustrated in Fig.~\ref{fig:region_interest_highlight}. This region corresponds to a grid which is subdivided into cells. Each grid represents the atmospheric state of the region of interest at a given hour on an arbitrary day. 
The region of interest covers a set of gauge stations, mainly located in Rio de Janeiro metropolitan area (blue-shaded area comprising three cells inside the region of interest). A surface station belongs to a cell if its location (latitude and longitude) falls within the cell’s spatial boundaries. In this region, there are 83 rain gauges from the Sirenes system, 33 rain gauges from the AlertaRio system, and 19 meteorological stations from INMET (National Institute of Meteorology). Black dots correspond to locations in which we have data from both ERA5 ans GFS.

\subsection{Dataset Building}
\label{sec:dataset:building}

\begin{figure}[htb]
    \centering
    \begin{tikzpicture}
        \node at (-1.8,0) {\Huge (};
        \node at (5.8,0) {\Huge )};
        
        \node at (0,0) {\includegraphics[width=3.5cm]{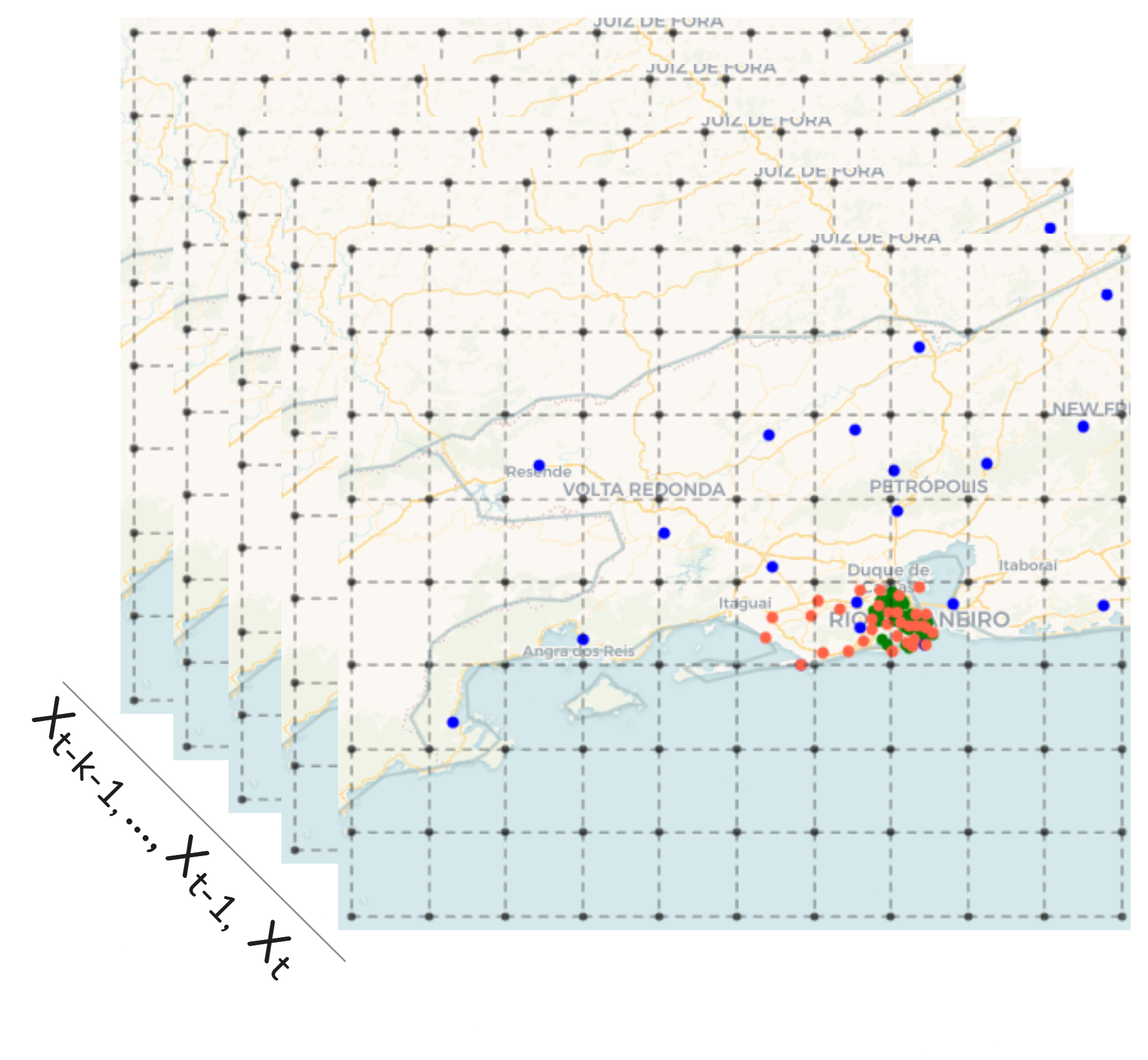}};
        
        \node at (2.0,0) {\Large ,};
        
        \node at (3.8,0) {\includegraphics[width=3.5cm]{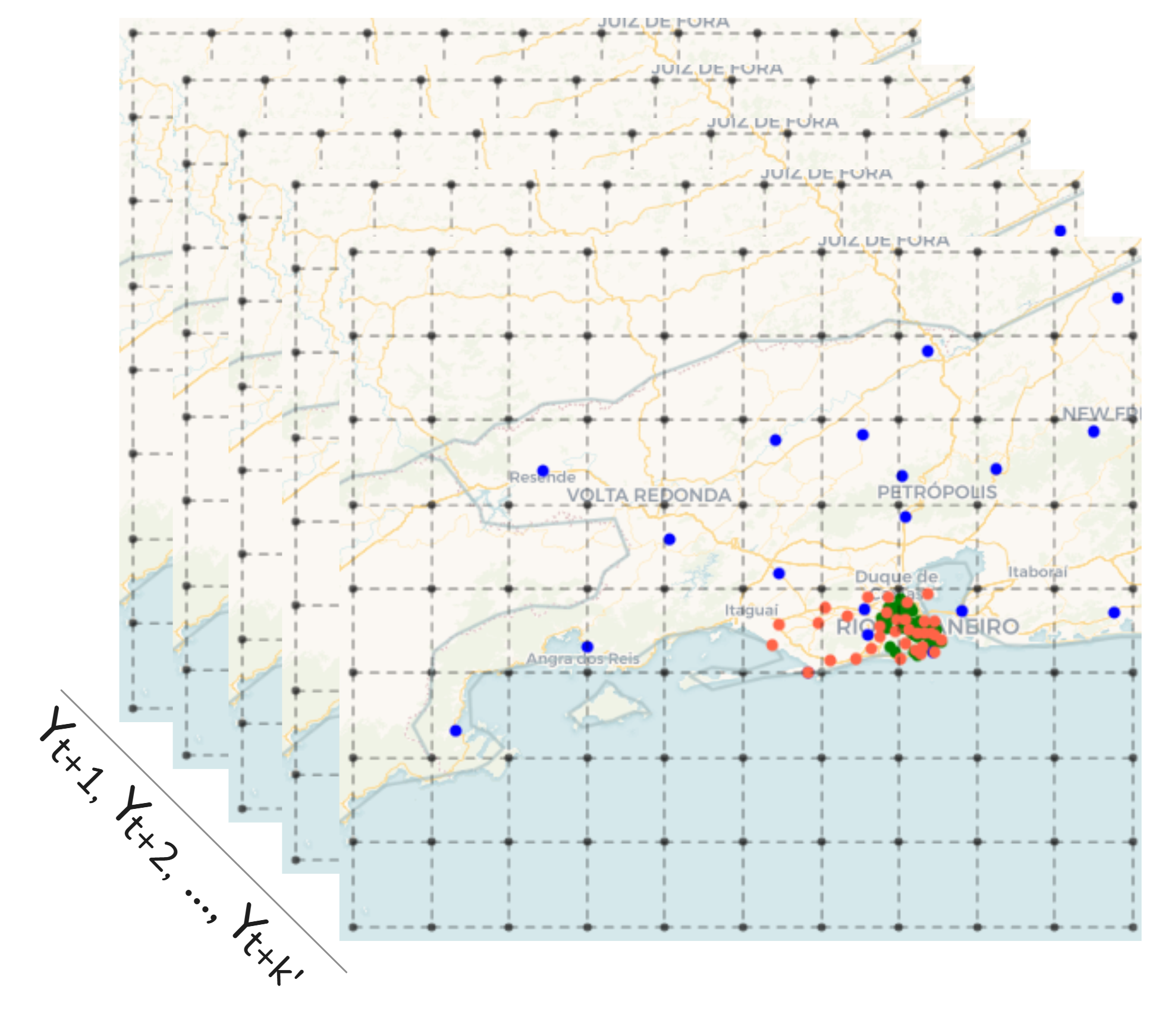}};
    \end{tikzpicture}
    \caption{An example $(X,Y)$ is a pair of grid sequences, in which $X = X_{t-k-1}, \dots, X_i, \dots, X_{t-1}, X_{t}$ is the grid sequence of input features, and $Y = Y_{t+1}, Y_{t+2}, \dots, Y_j, \dots, Y_{t+k'}$ is the grid sequence of precipitation measurements (\( t \) represents the time step). Each cell in \( X_i \) contains measurements of precipitation, temperature, humidity, pressure, wind speed, and wind direction. Each cell in \( Y_j \) contains precipitation measurements.}
    \label{fig:grid_input_output}
\end{figure}
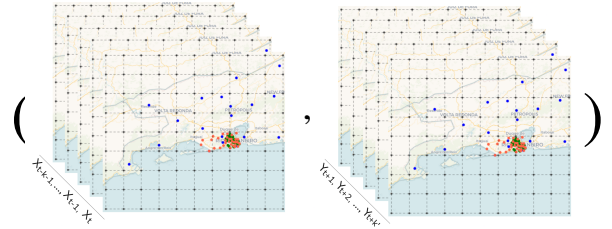

We now detail the procedure to build the dataset to be used for training the spatiotemporal nowcasting model. This procedure integrates precipitation data from multiple stations into a gridded structure, with the purpose of obtaining spatiotemporal grids that incorporate the observed station measurements, ensuring the resulting dataset better reflects real precipitation conditions. A single example of the resulting dataset consists of a set of hourly grids as input and a set of grids with precipitation measurements as the model's expected output, following the sequence-to-sequence approach~\cite{s2s}. Fig.~\ref{fig:grid_input_output} pictorially represents a single example $(X,Y)$ where the grid sequence $X$ corresponds to features, and the grid sequence $Y$ corresponds to the expected precipitation measurements. Here, $k$ corresponds to the lookback window, while $k'$ denotes the forecasting horizon.

Each cell in a grid sequence $X$ holds the values of 19 variables, which are the input features. These include surface precipitation measurements and 6 meteorological variables measured at three atmospheric standard pressure levels, covering lower, middle and higher troposphere: 200, 700, and 1000 hPa. These pressure levels were selected by consulting a meteorologist. For each pressure level, the variables are temperature, relative humidity, U (zonal) and V (meridional) wind components, wind speed, and vertical wind velocity. The cells in the target spatiotemporal grid sequence $Y$ contain only precipitation measurements.


For each cell, there are two cases: (1) the cell contains at least one gauge station, and (2) the cell does not contain any gauge stations. In case (1), we get the maximum precipitation value among the precipitation measurements recorded by the stations inside that cell. This value will serve as the \textit{target} for the given cell (in any grid sequence $Y$), or as the lagged precipitation value (in any grid sequence $X$). In case (2), as a fall-back mechanism, we set the maximum precipitation value among the four ERA5 grid points located at the corners of the cell as the \textit{target}. In case (1), the maximum value is chosen to emphasize extreme precipitation events in the data distribution. 

Algorithm~\ref{alg:max_precipitation_generic} presents the pseudocode for determining the maximum precipitation value measured by a given system of stations $S$ in a cell located at coordinates \((i, j)\). The function \textsc{GetStations} (line~\ref{GetStations}) returns the set of stations from the system $S$ that were operating at timestamp $t$, and $P(s,t)$ (line~\ref{GetMax}) is the precipitation value measured by station $s$ at timestamp $t$.

    
    
    
    

\begin{algorithm}[htb]
    \caption{\textsc{FindMaxPrecip}: Find the maximum precipitation for station system \( S \) in cell \( (i, j) \) at timestamp $t$.}
    \label{alg:max_precipitation_generic}
    \begin{algorithmic}[1]
    \Require Cell location \( (i, j) \), station system \( S \), timestamp \( t \)
    \Ensure The maximum precipitation value \( p \) for cell \( (i, j) \) $\in$ \( S \) at time \( t \)
    
    \State $stations \gets \textsc{GetStations}(S, i, j, t)$ \label{GetStations}
    \State $p \gets \max\left[P(s,t) \text{ for } s \in stations\right]$ \label{GetMax}
    
    \State \Return $p$
    \end{algorithmic}
\end{algorithm}

Algorithm~\ref{alg:max_precipitation_generic} can be used to compute the maximum precipitation values across a non-empty set of station systems. In fact, Algorithm~\ref{alg:find_max_precipitation_across_systems} uses it to compute the maximum precipitation value considering all the station systems we use in this paper. In our computational experiments, we consider all the possible combinations of station systems (see Table~\ref{table:datasets}).

\begin{algorithm}[htb]
    \caption{\textsc{FindMaxPrecipitationAcrossSystems}: Find the maximum precipitation across multiple station systems for cell \( (i, j) \) at timestamp \( t \).}
    \label{alg:find_max_precipitation_across_systems}
    \begin{algorithmic}[1]
    \Require Cell location \( (i, j) \), timestamp \( t \)
    \Ensure The maximum precipitation value \( p \) for cell \( (i, j) \) across all systems at time \( t \)
    
    \State $tp\_sirenes \gets \textsc{FindMaxPrecip}(i, j, \text{Sirenes}, t)$
    \State $tp\_inmet \gets \textsc{FindMaxPrecip}(i, j, \text{INMET}, t)$
    \State $tp\_alertario \gets \textsc{FindMaxPrecip}(i, j, \text{AlertaRio}, t)$
    
    \State $p \gets \max (tp\_sirenes, tp\_inmet, tp\_alertario)$
    
    \State \Return $p$
    \end{algorithmic}
\end{algorithm}

One relevant technical detail we had to consider is that each system of gauge stations operates at a specific temporal resolution. In particular, the Sirenes and AlertaRio systems provide precipitation observations every fifteen minutes, whereas INMET records observations only once per hour. To unify the temporal resolution across stations, we aggregate the values from the station system with the higher temporal resolution to match that of the lower-resolution station. For instance, we sum the precipitation measurements recorded by a Sirenes station at 14h15, 14h30, 14h45, and 15h00 to obtain its accumulated precipitation at 15h00. 

We build each example $\left(X, Y\right)$ using the sliding window technique. For example, considering a look-back window of size 5, a forecasting horizon of 5, and given a sequence of 11 time steps \( T_1, T_2, \dots, T_{11} \) (each step corresponding to a grid of observations in the region of interest), we set grids from \( T_1, \dots, T_5 \) as $X$, and grids from \( T_6, \dots, T_{10} \) as $Y$. Then, we shift the sliding window one position and apply the same procedure up to \( T_{11} \). This results in a total of two examples. In general, the number of examples $n$ can be computed as \( n = \text{\#total\_timesteps} - (\text{\#lookback\_window} + \text{\#forecast\_horizon}) + 1 \).

We applied the above procedure to historical data from 1st of January 2011 to 31st of October 2024. This resulted in a total of 120,529 time steps (hours), i.e., 120,529 grids. We disregarded the dry months (which, for the chosen region of interest, correspond to June, July, and August), hence this total reduced to 89,617. When ignoring a specific month, we accounted for temporal discontinuity to ensure data consistency. For instance, consider the grids for time steps \( T_1, T_2, \dots, T_{11} \), where \( T_1, \dots, T_5 \) belong to May and \( T_6, \dots, T_{11} \) belong to September, as the intermediate months were ignored. In this case, it is not possible to construct a single sample \( T_4, T_5, T_6, T_7, T_8 \) that spans both months. However, it remains feasible to maintain separate samples \( T_1, T_2, T_3, T_4, T_5 \) and \( T_6, T_7, T_8, T_9, T_{10} \) correctly. Consequently, the total number of samples is reduced when excluding the dry months, as expected. At the end of the processing, the dataset contains a total of 89,482 time steps.

Thus, defining the total number of time steps as \( n = 89482 \), the look-back window size as \( 5 \), the forecasting horizon as \( 5 \), the number of latitudes as 9, the number of longitudes as 11, and the number of channels as 19 for the feature channels and 1 for the target channel, the tensor shapes for the features and target are given by \( (89482, 5, 9, 11, 19) \) and \( (89482, 5, 9, 11, 1) \), respectively. 


\subsection{Model Building and Evaluation}
\label{sec:model_building}

After executing the procedure detailed in Section~\ref{sec:dataset:building} to construct the spatiotemporal dataset, we are able to train the precipitation nowcasting model using the STConvS2S deep learning architecture, which was designed to capture spatial and temporal data dependencies using only convolutional layers~\cite{CASTRO2021285}. This Deep Learning architecture consists of the temporal block, spatial block, and temporal generator block. The temporal block learns the temporal representation while preserving the temporal order (causal constraint). The spatial block extracts spatial features. The temporal generator block enables the generation of output sequences that are greater than or equal to the length of the input sequence.


The dataset is divided (in accordance with chronological order) into 60\% for training (January 2011 to February 2019), 20\% for validation (February 2019 to December 2021), and 20\% for testing (December 2021 to October 2024). The model error is computed using a weighted MAE loss, emphasizing rarer precipitation levels. In addition to evaluating the error, we assess the models using the confusion matrix, mean bias, and the F1-Score for four precipitation intensity levels: weak, moderate, heavy, and extreme. These levels are defined as follows: \( [0, 5) \) mm/h = weak, \( [5, 25) \) mm/h = moderate, \( [25, 50) \) mm/h = heavy, and \( [50, \infty) \) mm/h = extreme.

We observe (see Fig.~\ref{fig:region_interest_highlight}) that data fusion is effectively achieved only in the cells with a higher concentration of surface stations, which are located in the capital of Rio de Janeiro. Therefore, for model evaluation, we consider the assessment metrics within these cells, where the density of surface stations is highest. This approach allows us to specifically evaluate regions where heavy and extreme precipitation measurements are observed from the surface station sensors. For model training, however, we consider all cells within the region of interest.

\section{Experiments}
\label{sec:experiments}

In this Section, we present the computational experiments conducted to evaluate our spatiotemporal fusion approach. Section~\ref{sec:dataset_versions} describes the eight dataset versions created by combining different surface station networks with ERA5 reanalysis data. Section~\ref{sec:exploratory_analysis} presents an exploratory analysis of our datasets, including visualization of precipitation distributions, sanity checks comparing different data sources, and verification that our fusion approach correctly represents extreme precipitation events. Section~\ref{sec:comparative_experiments} evaluates the performance of our models using MAE, bias, confusion matrices and F1-scores for the different precipitation levels. Furthermore, we present an experiment integrating the GFS model with surface station observations in the inference phase, for lead times from 1 to 5 hours.

\subsection{Dataset Versions}
\label{sec:dataset_versions}

Using the procedure for dataset construction outlined in Section~\ref{sec:dataset:building}, we combined different available gauge station systems to produce eight dataset versions, which are summarized in Table~\ref{table:datasets}. The \texttt{ERA5} dataset does not include any surface station data; \texttt{ERA5+I} includes only INMET data combined with ERA5; similarly, \texttt{ERA5+A}, \texttt{ERA5+S}, \texttt{ERA5+IA}, \texttt{ERA5+SA}, \texttt{ERA5+SI}, and \texttt{ERA5+SIA} correspond to combinations of ERA5 with various gauge station systems, where the last dataset integrates INMET, Sirenes, and AlertaRio systems with ERA5.

\begin{table}[htb]
    \centering
    \renewcommand{\arraystretch}{1.4}
    \setlength{\tabcolsep}{10pt}
    \caption{Datasets versions used in the computational experiments.}
    \begin{tabular}{lccc}
    \toprule
     Dataset & Sirenes & INMET & AlertaRio \\
    \midrule
    \texttt{ERA5} &  &  &  \\ 
    \texttt{ERA5+S} & \checkmark &  &  \\ 
    \texttt{ERA5+I} &  & \checkmark &  \\ 
    \texttt{ERA5+A} &  &  & \checkmark \\ 
    \texttt{ERA5+SI} & \checkmark & \checkmark &  \\ 
    \texttt{ERA5+SA} & \checkmark &  & \checkmark \\ 
    \texttt{ERA5+IA} &  & \checkmark & \checkmark \\ 
    \texttt{ERA5+SIA} & \checkmark & \checkmark & \checkmark \\ 
    \bottomrule
    \end{tabular}
    \label{table:datasets}
\end{table}

\subsection{Exploratory Analysis}
\label{sec:exploratory_analysis}

\begin{figure}[htb]
    \centering
    \includegraphics[width=0.48\textwidth]{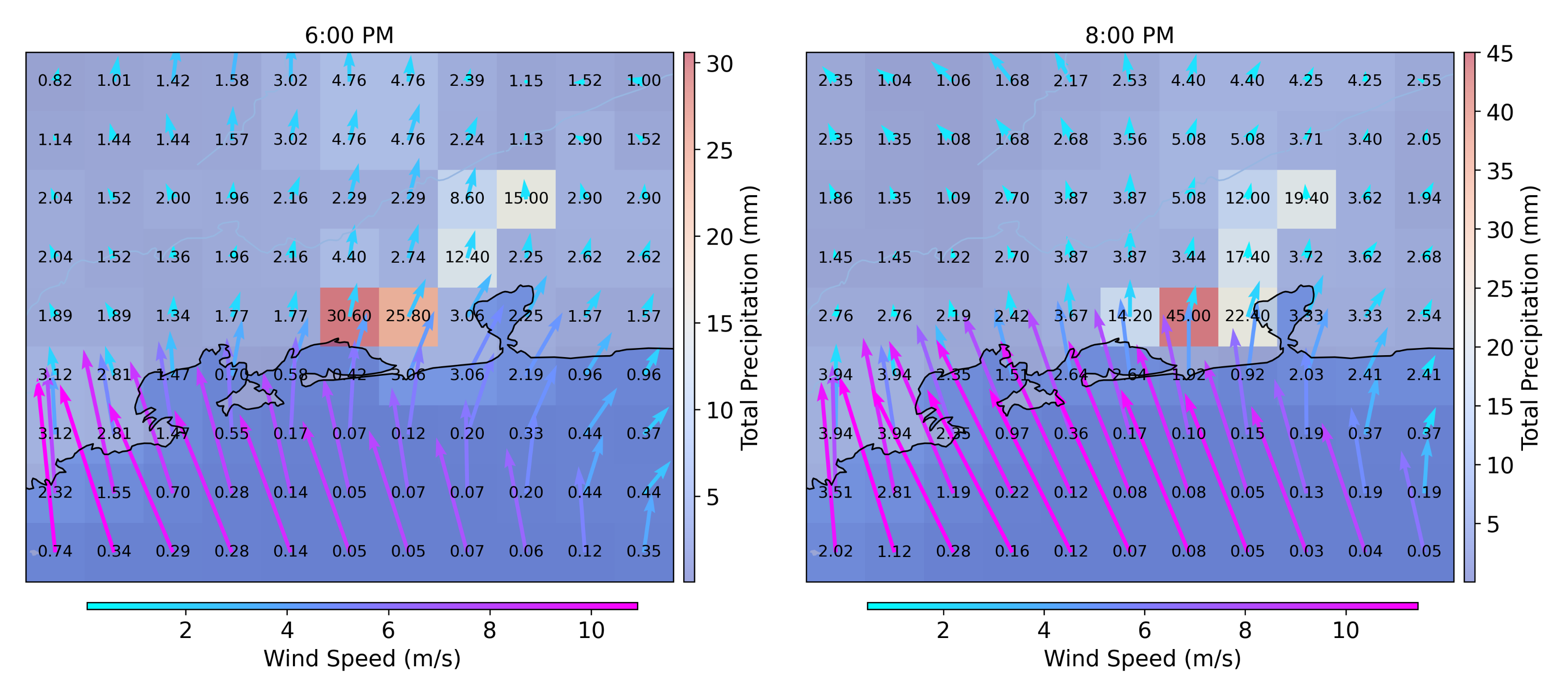}
    \caption{Precipitation heatmaps for January 13, 2024, at 6:00 PM and 8:00 PM.}
    \label{fig:analise_exploratoria}
\end{figure}

We conducted exploratory analyses to verify whether the algorithm presents consistent data on days with extreme rainfall events, as recorded in the knowledge base, and to visualize the algorithm's output. Fig.~\ref{fig:analise_exploratoria} presents two precipitation heatmaps for January 13, 2024\footnote{G1 News: \href{https://g1.globo.com/rj/rio-de-janeiro/noticia/2024/01/13/chuvas-no-rj-tem-ruas-alagadas-quedas-de-arvores-e-problemas-no-transito.ghtml}{Heavy rain in Rio de Janeiro causes flooded streets, fallen trees, and traffic issues}}, at 18h00 and 20h00. As a result of the data fusion, we observe the presence of extreme rainfall in the Rio de Janeiro region, with values exceeding 50 mm/h. Additionally, we note the uniformity and consistency of the data, a characteristic of the ERA5 model. Furthermore, we observe the consistency of the data with the knowledge base. We performed this analysis for other days with extreme precipitation events, as well as days with no rain.

\begin{figure}[htb]
    \centering
    \includegraphics[width=0.45\textwidth]{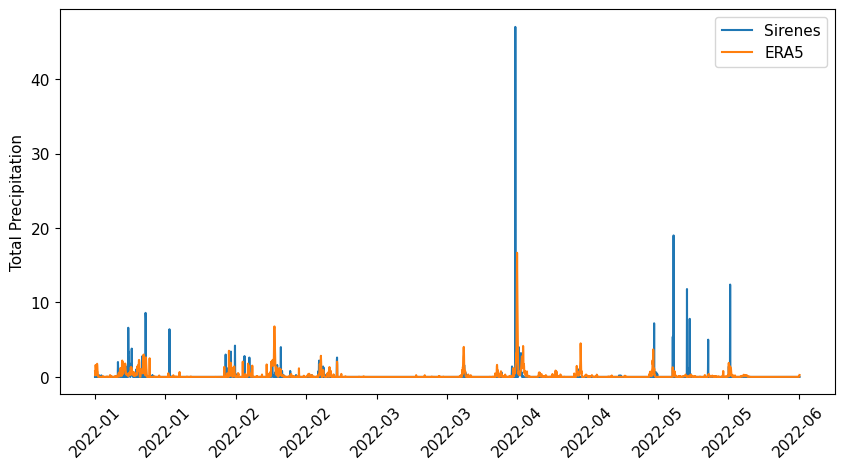}
    \caption{Sanity check to compare the precipitation measurements from the 'Pavão-Pavãozinho 2' station of the Sirenes system and the ERA5 reanalysis model between January 2022 and June 2022.}
    \label{fig:Total_Precipitation_values_sanity_check}
\end{figure}

We performed sanity checks to verify the precipitation distributions from the data sources we used, including reanalysis models, numerical models, and surface stations. Fig.~\ref{fig:Total_Precipitation_values_sanity_check} presents a sanity check comparing the precipitation measurements from the 'Pavão-Pavãozinho 2' station of the Sirenes system and the ERA5 reanalysis model between January 2022 and June 2022. We observe that the precipitation data from both data sources present a similar pattern of rainfall occurrence sparsity. However, the precipitation measurements coming from the Sirenes station are consistently greater than the corresponding values from the ERA5 model.

\begin{figure}[htb]
    \centering
    \includegraphics[width=0.45\textwidth]{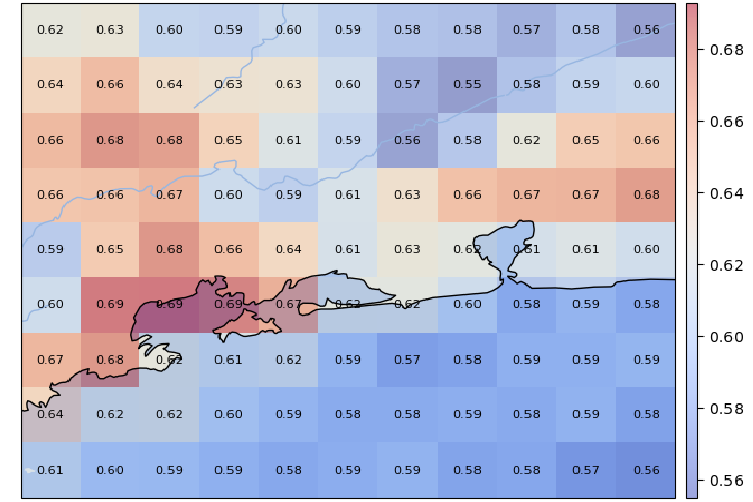}
    \caption{Spearman correlation for each cell in the spatiotemporal grid from January 2023 to December 2024. Each grid cell represents the Spearman correlation between GFS and ERA5 from January 2023 to December 2024.}
    \label{fig:spearmen_correlation}
\end{figure}

Fig.~\ref{fig:spearmen_correlation} shows the Spearman correlation between the GFS and ERA5 models for each cell in the spatiotemporal grid from January 2023 to December 2024. We observe that the Spearman correlation is greater than 0.5 for all cells, indicating a positive correlation between the data sources.

\begin{figure}[htb]
    \centering
    \includegraphics[width=0.45\textwidth]{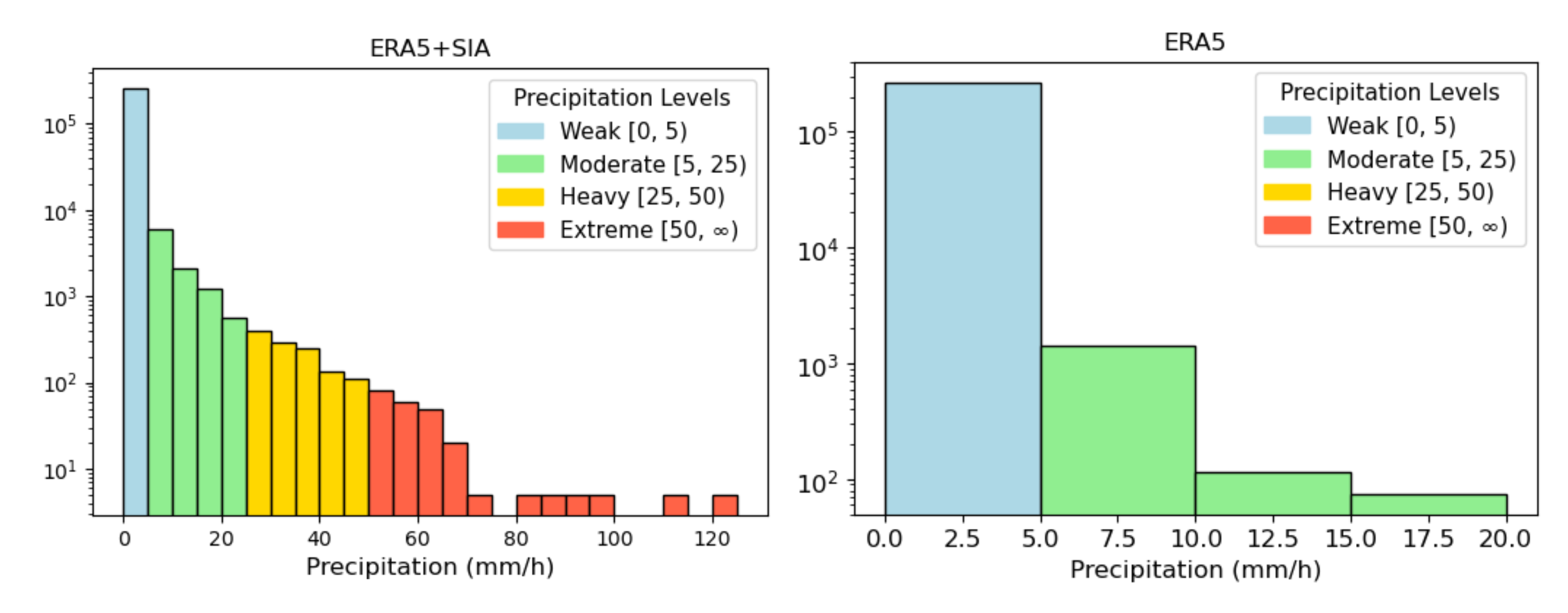}
    \caption{Log-scale histogram of precipitation measurements for the precipitation levels. The log scale was used to improve visualization due to data imbalance, i.e., the majority of observations fall within the ``Weak'' level. We observe that in the \texttt{ERA5} dataset, there are no precipitation observations for the ``Heavy'' and ``Extreme'' levels. In contrast, in the \texttt{ERA5+SIA} dataset, all four precipitation levels are present.}
    \label{fig:distribuition-test-dataset}
\end{figure}

When evaluating one of the eight test datasets, we are interested in ensuring that the distribution includes the four levels of precipitation defined as ``Weak'', ``Moderate'', ``Heavy'', and ``Extreme''. In Fig.~\ref{fig:distribuition-test-dataset}, we observe that the \texttt{\texttt{ERA5+SIA}} test dataset contains all precipitation levels, whereas the \texttt{ERA5} dataset lacks the ``Heavy'' and ``Extreme'' precipitation levels.

\subsection{Comparative Experiments}
\label{sec:comparative_experiments}



We present in Table~\ref{table:mae} and Table~\ref{table:bias} the MAE error and Bias for the eight experiments conducted, respectively. We observe that the \texttt{ERA5+SIA} model exhibits an MAE error of $41.8832$ mm/h for the extreme level. For the majority class level ``Weak'', the error is close to $0$ mm/h for the same model. We also observe that the model's bias assumes negative values, especially for the ``Extreme'' class, indicating that the model tends to underestimate the precipitation level.

\begin{table}[htb]
    \renewcommand{\arraystretch}{1.4}
    \setlength{\tabcolsep}{10pt}
    \centering
    \caption{MAE error for the eight experiments conducted with STConvS2S.}
    \begin{tabular}{lrrrr}
    \toprule
     & [0-5) & [5-25) & [25-50) & [50-inf) \\
    \midrule
    \texttt{ERA5} & \textbf{0.1217} & \textbf{2.8232} & nan & nan \\
    \texttt{ERA5+S} & 0.3499 & 5.1270 & 21.4554 & 50.3021 \\
    \texttt{ERA5+I} & 0.1484 & 7.7168 & 31.7476 & 59.5871 \\
    \texttt{ERA5+A} & 0.3162 & 7.2699 & 23.8799 & 44.4496 \\
    \texttt{ERA5+SI} & 0.2702 & 9.4626 & 34.6405 & 64.2387 \\
    \texttt{ERA5+SA} & 0.4267 & 6.3092 & 20.5067 & 43.1067 \\
    \texttt{ERA5+IA} & 0.3524 & 6.8721 & 23.1909 & 44.5691 \\
    \texttt{ERA5+SIA} & 0.4392 & 6.3858 & \textbf{19.5797} & \textbf{41.8832} \\
    \bottomrule
    \end{tabular}
    \label{table:mae}
\end{table}

\begin{table}[htb]
    \renewcommand{\arraystretch}{1.4}
    \setlength{\tabcolsep}{10pt}
    \centering
    \caption{Bias for the eight experiments performed with STConvS2S.}
    \begin{tabular}{lrrrr}
    \toprule
     & [0-5) & [5-25) & [25-50) & [50-inf) \\
    \midrule
    \texttt{ERA5} & \textbf{-0.0332} & -2.3567 & nan & nan \\
    \texttt{ERA5+S} & 0.2572 & \textbf{-1.1296} & -21.4554 & -50.3021 \\
    \texttt{ERA5+I} & 0.0524 & -7.4414 & -31.7476 & -59.5871 \\
    \texttt{ERA5+A} & 0.1829 & -5.5719 & -23.3327 & -44.3141 \\
    \texttt{ERA5+SI} & -0.2693 & -9.4626 & -34.6405 & -64.2387 \\
    \texttt{ERA5+SA} & 0.2443 & -1.7719 & -18.7895 & -43.0711 \\
    \texttt{ERA5+IA} & 0.2021 & -4.4897 & -22.5589 & -44.5691 \\
    \texttt{ERA5+SIA} & 0.2487 & -1.5463 & \textbf{-18.0461} & \textbf{-41.5990} \\
    \bottomrule
    \end{tabular}
    \label{table:bias}
\end{table}

In the experiments we conducted, we evaluated lead times of 1, 2, 3, 4, and up to 5 hours ahead. Table~\ref{table:confusion_matrix} presents the Confusion Matrix for a lead time of 1 hour. The rows correspond to the expected/actual classes, while the columns represent the predicted classes. Specifically, for the extreme precipitation level, we observe that although the model occasionally misclassifies ``Extreme'' rainfall as ``Weak'', it predominantly assigns it to the ``Moderate'', ``Heavy'', or ``Extreme'' categories.

\begin{table}[htb]
    \renewcommand{\arraystretch}{1.4}
    \setlength{\tabcolsep}{10pt}
    \centering
    \caption{Confusion Matrix for the 1-hour Lead Time of the \texttt{ERA5+SIA} Model.}
    \begin{tabular}{lrrrr}
    \toprule
    & [0-5) & [5-25) & [25-50) & [50-inf) \\
    \midrule
    \textnormal{[0-5)} & 49959 & 1416 & 40 & 2 \\
    \textnormal{[5-25)} & 803 & 1090 & 88 & 7 \\
    \textnormal{[25-50)} & 48 & 143 & 43 & 3 \\
    \textnormal{[50-inf)} & 2 & 31 & 15 & 1 \\
    \bottomrule
    \end{tabular}
    \label{table:confusion_matrix}
\end{table}

\begin{figure}[htb]
    \centering
    \includegraphics[width=0.48\textwidth]{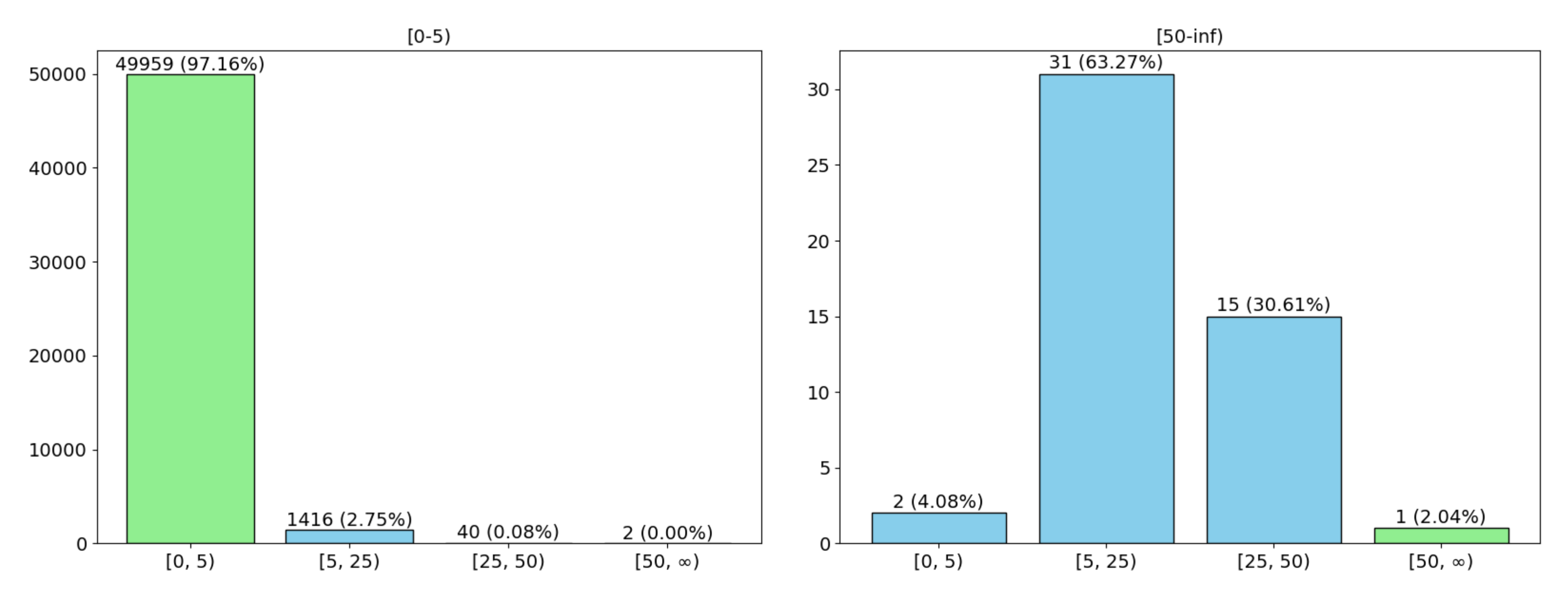}
    \caption{Histogram presenting a pictorial view of the confusion matrix for the 1-hour lead time of the \texttt{ERA5+SIA} model for the levels ``Weak'' and ``Extreme''. We observe that for the extreme level, the model correctly predicted 1 instance of extreme rainfall (2.04\%). Also, 2 (4.08\%), 31 (63.27\%), and 15 (30.61\%) instances of extreme rainfall were incorrectly predicted as ``Weak'', ``Moderate'', and ``Heavy'' levels, respectively. While the model underestimated the intensity of extreme events, it recognized 93.88\% of them as significant precipitation (either ``Moderate'' or ``Heavy''). A similar observation holds for the weak level; for instance, 99\% of weak rainfall instances were correctly predicted.}
    \label{fig:confusion_matrix_extreme}
\end{figure}

We highlight in Fig.~\ref{fig:confusion_matrix_extreme} the ``Weak'' and ``Extreme'' levels of the Confusion Matrix for the 1-hour lead time of the \texttt{ERA5+SIA} model. The taller the bar corresponding to a given level, the better the model classifies precipitation. We observe that most observations at the ``Extreme'' level are predicted as ``Heavy'', indicating that the model is biased towards underestimating precipitation levels.

Once we have the confusion matrix for each model, we compute the F1-Score of the experiments by precipitation level. Table~\ref{table:f1-score-1-lead-time} presents the F1-Score for each model at a 1-hour lead time ($T+1$). We observe that the \texttt{ERA5+SIA} model achieves an F1-Score of 0.2033 for the heavy level. For the majority class ``Weak'', the F1-Score is approximately 0.9774 for the same model. Table~\ref{table:f1-score-each-lead-time} presents the F1-Score for the \texttt{ERA5+SIA} model at each lead time individually. Table~\ref{table:f1-score-all-lead-times} presents the F1-Score for each model at lead times up to 5 hours, i.e., evaluating the combined lead times rather than individually. In Table~\ref{table:f1-score-each-lead-time}, we observe that the model's performance gradually decreases as the lead time increases. For example, this degradation in F1-scores is evident for heavy events, dropping from 0.2033 at T+1 to 0.0265 at T+5. This behavior is expected and inherent to precipitation nowcasting, as longer-term predictions are increasingly challenging, a characteristic also observed by \cite{rainfall2021} in their precipitation forecasting experiments.

\begin{table}[htb]
    \renewcommand{\arraystretch}{1.4}
    \setlength{\tabcolsep}{10pt}
    \centering
    \caption{F1-Score for the eight experiments conducted with STConvS2S at a 1-hour lead time (T+1 in the temporal dimension).}
    \begin{tabular}{lrrrr}
    \toprule
    & [0-5) & [5-25) & [25-50) & [50-inf) \\
    \midrule
    \texttt{ERA5} & \textbf{0.9973} & 0.4387 & 0.0000 & 0.0000 \\
    \texttt{ERA5+S} & 0.9867 & 0.3450 & 0.0294 & 0.0000 \\
    \texttt{ERA5+I} & 0.9946 & 0.1869 & 0.0000 & 0.0000 \\
    \texttt{ERA5+A} & 0.9841 & 0.3155 & 0.1782 & \textbf{0.0588} \\
    \texttt{ERA5+SI} & 0.9903 & 0.0000 & 0.0000 & 0.0000 \\
    \texttt{ERA5+SA} & 0.9788 & 0.4532 & 0.1652 & 0.0351 \\
    \texttt{ERA5+IA} & 0.9810 & 0.3840 & 0.0846 & 0.0000 \\
    \texttt{ERA5+SIA} & 0.9774 & \textbf{0.4670} & \textbf{0.2033} & 0.0323 \\
    \bottomrule
    \end{tabular}
    \label{table:f1-score-1-lead-time}
\end{table}

\begin{table}[htb]
    \renewcommand{\arraystretch}{1.4}
    \setlength{\tabcolsep}{10pt}
    \centering
    \caption{F1-Score for the \texttt{ERA5+SIA} model, evaluated at each lead time (T+1 to T+5).}
    \begin{tabular}{lrrrr}
    \toprule
    & [0-5) & [5-25) & [25-50) & [50-inf) \\
    \midrule

    T+1 & \textbf{0.9774} & \textbf{0.4670} & 0.2033 & 0.0323 \\
    T+2 & 0.9701 & 0.3650 & \textbf{0.2042} & 0.0614 \\
    T+3 & 0.9649 & 0.2946 & 0.1319 & \textbf{0.0663} \\
    T+4 & 0.9671 & 0.3031 & 0.1331 & 0.0339 \\
    T+5 & 0.9710 & 0.2872 & 0.0265 & 0.0000 \\
    \bottomrule
    \end{tabular}
    \label{table:f1-score-each-lead-time}
\end{table}

\begin{table}[htb]
    \renewcommand{\arraystretch}{1.4}
    \setlength{\tabcolsep}{10pt}
    \centering
    \caption{F1-Score for the STConvS2S model, evaluated with combined lead times (T+1 to T+5).}
    \begin{tabular}{lrrrr}
    \toprule
    & [0-5) & [5-25) & [25-50) & [50-inf) \\
    \midrule
    \texttt{ERA5} & \textbf{0.9970} & 0.3106 & 0.0000 & 0.0000 \\
    \texttt{ERA5+S} & 0.9799 & 0.2019 & 0.0742 & 0.0115 \\
    \texttt{ERA5+I} & 0.9938 & 0.1431 & 0.0000 & 0.0000 \\
    \texttt{ERA5+A} & 0.9810 & 0.2322 & 0.1111 & 0.0221 \\
    \texttt{ERA5+SI} & 0.9903 & 0.0000 & 0.0000 & 0.0000 \\
    \texttt{ERA5+SA} & 0.9709 & 0.3259 & 0.1176 & 0.0441 \\
    \texttt{ERA5+IA} & 0.9778 & 0.2983 & 0.1291 & 0.0353 \\
    \texttt{ERA5+SIA} & 0.9701 & \textbf{0.3438} & \textbf{0.1489} & \textbf{0.0535} \\
    \bottomrule
    \end{tabular}
    \label{table:f1-score-all-lead-times}
\end{table}

\begin{figure}[htb]
    \centering
    \includegraphics[width=0.45\textwidth]{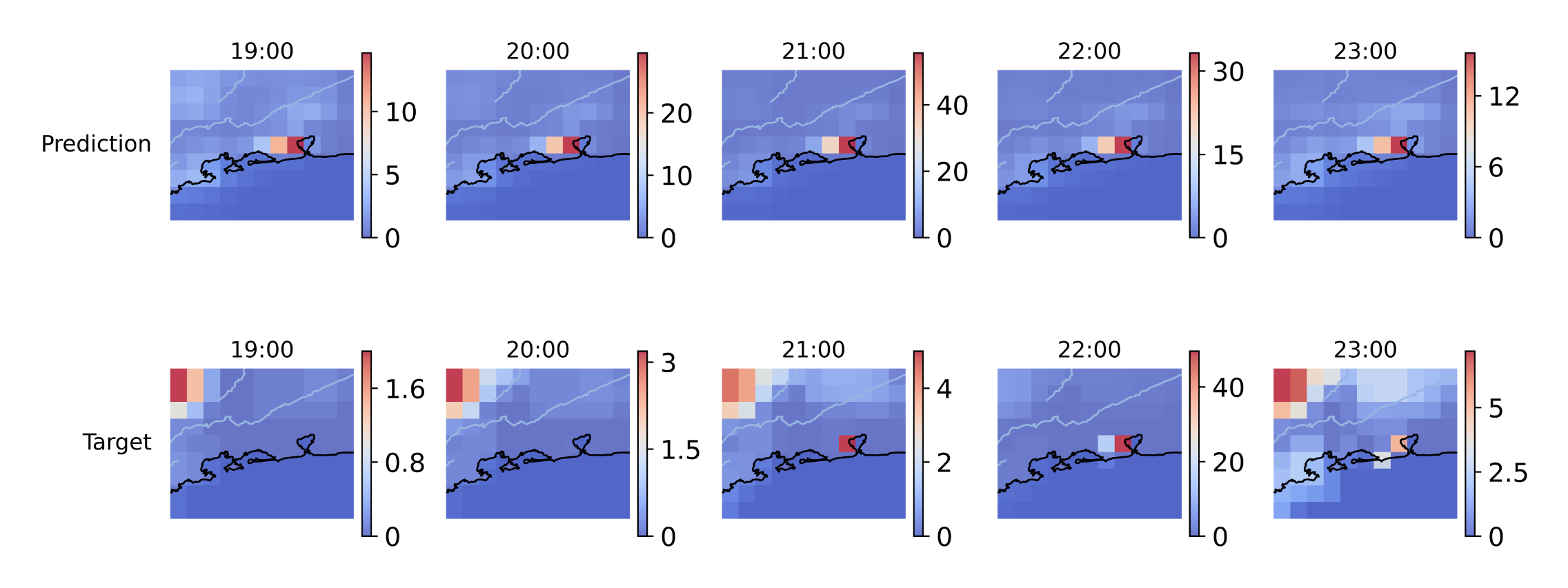}
    \caption{Precipitation Nowcast for December 20, 2024, \texttt{GFS+A} Dataset applied in the \texttt{ERA5+SIA} Model.}
    \label{fig:20-dec-2024-no-scale-normalize}
\end{figure}

After evaluating the models with the test set, we conducted inference-phase experiments. As described in Section~\ref{sec:methods}, we used the ERA5 reanalysis model during the training phase and the NWP model during the inference phase. In these integrated experiments, we combined the AlertaRio data with the GFS. Fig.~\ref{fig:20-dec-2024-no-scale-normalize} presents the precipitation forecast for December 20, 2024\footnote{G1 News: \href{https://g1.globo.com/rj/rio-de-janeiro/noticia/2024/12/20/chuva-no-rio-sexta-feira.ghtml}{Rio experienced heavy rain in several neighborhoods; hail was reported in Barra}}. We provided the model with the spatiotemporal grids at 14:00, 15:00, 16:00, 17:00, and 18:00 to forecast precipitation for 19:00, 20:00, 21:00, 22:00, and 23:00. We observed that the model produced predictions consistent with the expected input and output. Table~\ref{table:20-dec-2024} presents the Confusion Matrix for the same experiment. We observe that while the model successfully detects the occurrence of heavy precipitation events, the predicted precipitation measurements don't match the targets across lead times. For instance, at T4, where target precipitation suddenly spikes to 40 mm/h, the model predicts up to 30 mm/h, demonstrating its ability to identify significant events. These discrepancies stem from sudden temporal variations in precipitation measurements. Nevertheless, the model effectively identifies periods of increased rainfall intensity despite timing and quantity differences.

\begin{table}[htb]
    \renewcommand{\arraystretch}{1.4}
    \setlength{\tabcolsep}{10pt}
    \centering
    \caption{Confusion Matrix of the \texttt{GFS+A} Dataset applied in the \texttt{ERA5+SIA} Model: Precipitation Nowcast for December 20, 2024. The 15 classifications correspond to the 3 cells we evaluated, multiplied by the 5 lead times.}
    \begin{tabular}{lrrrr}
    \toprule
    & [0-5) & [5-25) & [25-50) & [50-inf) \\
    \midrule
    \textnormal{[0-5)} & 2 & 7 & 2 & 0 \\
    \textnormal{[5-25)} & 0 & 2 & 0 & 1 \\
    \textnormal{[25-50)} & 0 & 0 & 1 & 0 \\
    \textnormal{[50-inf)} & 0 & 0 & 0 & 0 \\
    \bottomrule
    \end{tabular}
    \label{table:20-dec-2024}
\end{table}
    




\section{Final Remarks}
\label{sec:conclusions}

In this study, we employ data fusion techniques to propose an algorithm that integrates pluviometric data from surface stations with data from the ERA5 reanalysis model. We incorporate data from the Sirenes, INMET, and AlertaRio station networks. Data fusion enhances data completeness, particularly by leveraging the maximum precipitation value among stations to emphasize extreme event forecasting. The algorithm constructs data within a spatiotemporal framework over a defined and uniform rectangular grid. This is feasible because we integrate the spatial resolution, uniformity, and completeness characteristics of the ERA5 reanalysis model with surface station observations to train the STConvS2S model.

We apply the proposed data fusion algorithm to construct different spatiotemporal datasets spanning from January 2011 to October 2024. We evaluate the results obtained using the STConvS2S model for precipitation nowcasting. In this evaluation, we employ metrics such as the Confusion Matrix and F1-Score to assess the performance of various computational experiments. The model, when integrated with the Sirenes, INMET, AlertaRio, and ERA5 stations (\texttt{ERA5+SIA}), achieved F1-Scores of 0.9774, 0.4670, 0.2033, and 0.0323 for the ``Weak'', ``Moderate'', ``Heavy'', and ``Extreme'' precipitation levels, respectively, for a lead time of 1 hour. Based on evaluation metrics, the model demonstrates promising results for forecasting at the next time step $T + 1$. We conducted preliminary experiments integrating the GFS numerical model and AlertaRio data during inference to assess the model’s performance in precipitation nowcasting.


As ongoing work, we are investigating approaches to overcome the spatial resolution limitation of 0.25 degrees inherited from the ERA5 and GFS models. We aim to achieve higher spatial resolutions in our spatiotemporal grids. In particular, we are evaluating the use of the ERA5-Land reanalysis model, which provides a spatial resolution of 0.1 degrees (approximately 10 km).


For future work, we will explore the use of the data fusion algorithm by integrating additional data sources beyond those presented, including stations from the National Center for Monitoring and Early Warning of Natural Disasters (CEMADEN). Furthermore, we will incorporate an increased number of AlertaRio stations, given its expansion. This will enhance data completeness and mitigate the issue where most grid cells in the region of interest lack coverage from surface stations.


Moreover, we will investigate bias correction techniques to adjust the model’s predictions, which tend to underestimate precipitation measurements, particularly for extreme events. Specifically for precipitation data, we will perform experiments using Log-transformation normalization. This normalization approach effectively reduces the right-skewness observed in the imbalanced precipitation data.

\section*{Code and data availability}

Code and instructions for replicating the experiments is available via GitHub (\url{https://github.com/AILAB-CEFET-RJ/fusion2025/}). The STConvS2S implementation is available as a GitHub repository (https://github.com/AILAB-CEFET-RJ/stconvs2s, last access: 18 December 2024). 

Gauge data from AlertaRio and INMET are available in their respective portals (\url{https://portal.inmet.gov.br}, \url{http://alertario.rio.rj.gov.br/download/dados-pluviometricos/}, last access: 31 December 2024). Gauge data from Sirenes system are available upon request. 

Model structures, trained models, and dataset versions can be found at \url{https://doi.org/10.5281/zenodo.14941806}.


\bibliography{references}

\begin{thebibliography}{16}
\providecommand{\natexlab}[1]{#1}
\providecommand{\url}[1]{\texttt{#1}}
\expandafter\ifx\csname urlstyle\endcsname\relax
  \providecommand{\doi}[1]{doi: #1}\else
  \providecommand{\doi}{doi: \begingroup \urlstyle{rm}\Url}\fi

\bibitem[Ahrens and Henson(2019)]{ahrens2019meteorology}
C.~D. Ahrens and R.~Henson.
\newblock \emph{Meteorology today: an introduction to weather, climate, and the
  environment}.
\newblock Cengage, twelfth edition edition, 2019.

\bibitem[Bleiholder and Naumann(2009)]{bleiholder2009}
J.~Bleiholder and F.~Naumann.
\newblock Data fusion.
\newblock \emph{ACM Comput. Surv.}, Jan. 2009.

\bibitem[Browning(1982)]{Browning1982}
K.~A. Browning.
\newblock \emph{Nowcasting}.
\newblock Academic Press, San Diego, CA, Aug. 1982.

\bibitem[Castro et~al.(2021)Castro, Souto, Ogasawara, Porto, and
  Bezerra]{CASTRO2021285}
R.~Castro, Y.~M. Souto, E.~Ogasawara, F.~Porto, and E.~Bezerra.
\newblock Stconvs2s: Spatiotemporal convolutional sequence to sequence network
  for weather forecasting.
\newblock \emph{Neurocomputing}, 426:\penalty0 285--298, 2021.

\bibitem[Choi and Cha(2021)]{rainFusion}
Y.~Choi and e.~a. Cha.
\newblock Rain-f: A fusion dataset for rainfall prediction using convolutional
  neural network.
\newblock pages 7145--7148, 2021.

\bibitem[Hersbach and Bell(2020)]{era5}
H.~Hersbach and e.~a. Bell.
\newblock The era5 global reanalysis.
\newblock pages 1999--2049, 2020.

\bibitem[Kim et~al.(2024)Kim, Jeong, and Kim]{KIM2024105529}
W.~Kim, C.-H. Jeong, and S.~Kim.
\newblock Improvements in deep learning-based precipitation nowcasting using
  major atmospheric factors with radar rain rate.
\newblock \emph{Computers \& Geosciences}, 2024.

\bibitem[Ko et~al.(2022)Ko, Lee, Hwang, Oh, Son, and Shin]{KO2022105072}
J.~Ko, K.~Lee, H.~Hwang, S.-G. Oh, S.-W. Son, and K.~Shin.
\newblock Effective training strategies for deep-learning-based precipitation
  nowcasting and estimation.
\newblock \emph{Computers \& Geosciences}, 2022.

\bibitem[{National Oceanic and Atmospheric Administration
  (NOAA)}(2022)]{noaa_gfs}
{National Oceanic and Atmospheric Administration (NOAA)}.
\newblock Noaa global forecast system (gfs), 2022.

\bibitem[Polifke(2019)]{polifke}
F.~P. d.~S. Polifke.
\newblock Projeto pesquisa operacional. relatório final (03/2019-06/2019),
  2019.

\bibitem[Reichstein et~al.(2019)Reichstein, Camps-Valls, Stevens, Jung,
  Denzler, Carvalhais, and Prabhat]{Reichstein2019195}
M.~Reichstein, G.~Camps-Valls, B.~Stevens, M.~Jung, J.~Denzler, N.~Carvalhais,
  and Prabhat.
\newblock Deep learning and process understanding for data-driven earth system
  science.
\newblock 2019.

\bibitem[Ronneberger et~al.()Ronneberger, Fischer, and Brox]{unet}
O.~Ronneberger, P.~Fischer, and T.~Brox.
\newblock U-net: Convolutional networks for biomedical image segmentation.

\bibitem[Srithagon et~al.(2021)Srithagon, Phisanbut, Piamsa-nga, and
  Piamsa-nga]{rainfall2021}
S.~Srithagon, N.~Phisanbut, N.~Piamsa-nga, and P.~Piamsa-nga.
\newblock Rainfall nowcasting based on neighboring rain gauge stations using
  learning machines.
\newblock 2021.

\bibitem[Sutskever et~al.(2014)Sutskever, Vinyals, and Le]{s2s}
I.~Sutskever, O.~Vinyals, and Q.~V. Le.
\newblock Sequence to sequence learning with neural networks.
\newblock Cambridge, MA, USA, 2014. MIT Press.

\bibitem[Tosiri and Kleawsirikul(2021)]{tosiri2021}
W.~Tosiri and e.~a. Kleawsirikul.
\newblock Precipitation nowcasting using deep learning on radar data augmented
  with satellite data.
\newblock 2021.

\bibitem[Wang and Coning(2017)]{wmo_nowcasting}
Y.~Wang and e.~a. Coning.
\newblock \emph{Guidelines for Nowcasting Techniques}.
\newblock 11 2017.

\end{thebibliography}
\bibliographystyle{abbrvnat}

\end{document}